\documentclass[10pt,twocolumn,letterpaper]{article}

\usepackage{cvpr}
\usepackage{times}
\usepackage{epsfig}
\usepackage{graphicx}
\usepackage{amsmath}
\usepackage{amssymb}
\usepackage{subfigure}
\usepackage{indentfirst}
\usepackage{bm}
\usepackage[marginal]{footmisc}
\usepackage{bbding}

\usepackage[pagebackref=true,breaklinks=true,letterpaper=true,colorlinks,bookmarks=false]{hyperref}

\cvprfinalcopy 

\def\cvprPaperID{2578} 

\ifcvprfinal\fi
\begin{document}

\title{Dual Temporal Memory Network for Efficient Video Object Segmentation}
\author{Kaihua Zhang$^1$, Long Wang$^1$, Dong Liu$^2$, Bo Liu$^3$, Qingshan Liu$^1$, Zhu Li$^4$\\
$^1$B-DAT and CICAEET, Nanjing University of Information
Science and Technology, Nanjing, China\\
 $^2$Netflix Inc.\\
$^3$JD Digits, Mountain View, CA, USA\\
$^4$ Dept. of CSEE, University of Missouri-Kansas City, Missouri 64110,
USA.\\
{\tt\small \{zhkhua,kfliubo\}@gmail.com}
}

\maketitle

\begin{abstract}
Video Object Segmentation (VOS) is typically formulated in a semi-supervised setting. Given the ground-truth segmentation mask on the first frame, the task of VOS is to track and segment the single or multiple objects of interests in the rest frames of the video at the pixel level. One of the fundamental challenges in VOS is how to make the most use of the temporal information to boost the performance. We present an end-to-end network which stores short- and long-term video sequence information preceding the current frame as the temporal memories to address the temporal modeling in VOS. Our network consists of two temporal sub-networks including a short-term memory sub-network and a long-term memory sub-network.
The short-term memory sub-network models the fine-grained spatial-temporal interactions between local regions across neighboring frames in video via a graph-based learning framework, which can well preserve the visual consistency of local regions over time. The long-term memory sub-network models the long-range evolution of object via a Simplified-Gated Recurrent Unit (S-GRU), making the segmentation be robust against occlusions and drift errors. In our experiments, we show that our proposed method achieves a favorable and competitive performance on three frequently-used VOS datasets, including DAVIS 2016, DAVIS 2017 and Youtube-VOS in terms of both speed and accuracy.
\end{abstract}


\section{Introduction}
\begin{figure}
\begin{center}
\begin{tabular}{c}
\includegraphics[width=1\linewidth]{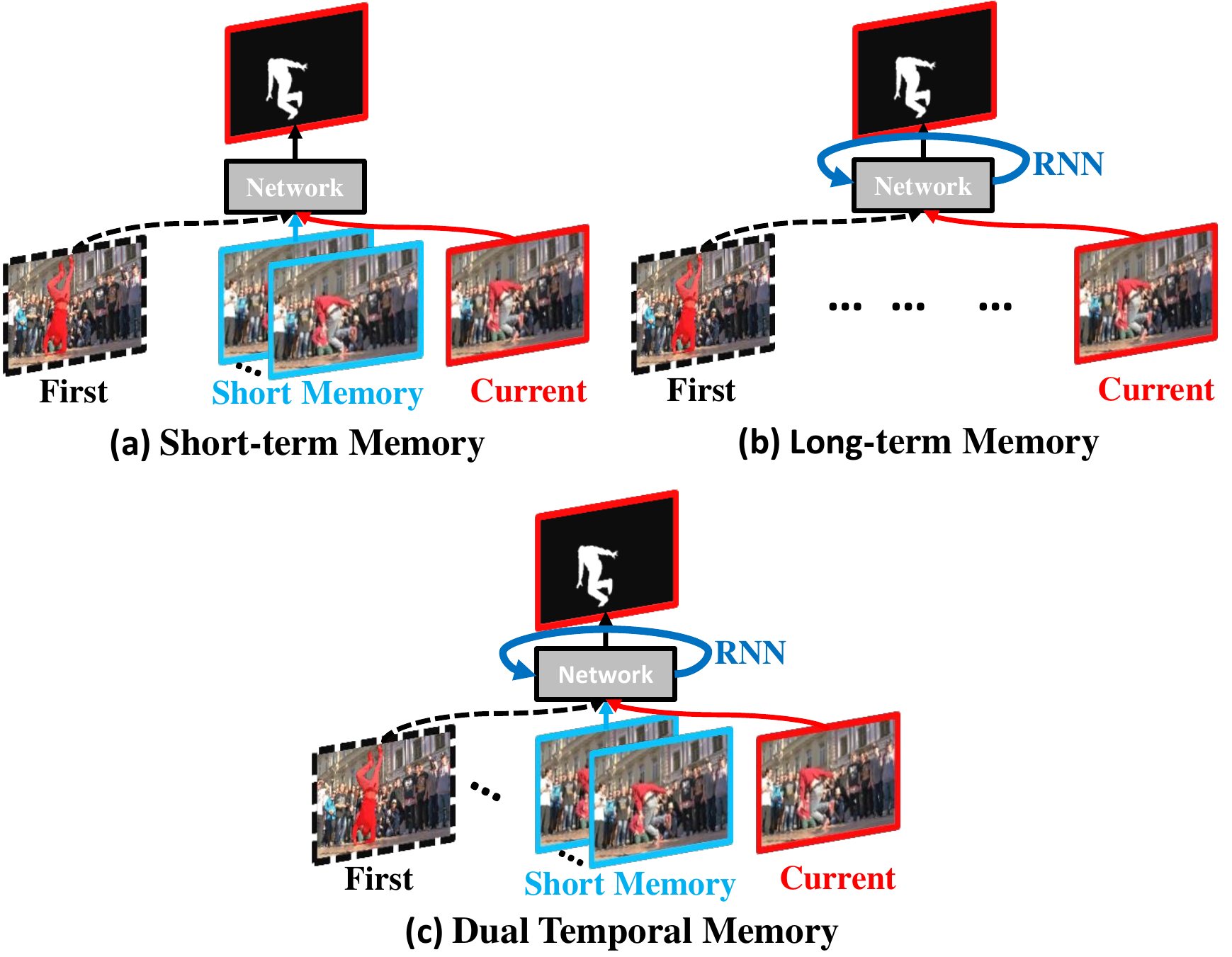}
\end{tabular}
\end{center}
   \caption{Previous methods~\cite{perazzi2017learning, khoreva2017lucid, cheng2017segflow, ventura2019rvos, xu2018youtube_s2s, tokmakov2017learning} capture temporal dependencies in a short-term or long-term video sequence for VOS (a,b). Our proposed method leverages both short- and long-term temporal information (c). RNN is short for Recurrent Neural Network.} 
\label{fig:ls}
\end{figure}
Video Object Segmentation (VOS) aims to separate the foreground objects from the backgrounds in all frames of a video sequence. The common approach casts the problem into a semi-supervised learning task, \textit{i.e.,} the segmentation ground truth of the target object in the first frame is provided and the goal is to infer the segmentation masks of the object in all other frames~\cite{Caelles2017one-shot, voigtlaender2017online, perazzi2017learning, xu2019mhp, yang2018efficient, wug2018fast,voigtlaender2019feelvos, ventura2019rvos, johnander2019generative}. Fast and accurate VOS methods are beneficial to many applications such as video editing~\cite{li2005video, wang2017selective}, object tracking~\cite{wang2019fast, yeo2017superpixel} and activity recognition~\cite{guo2013video}.
Modelling the inter-frame temporal correlation is one of the essential challenges in VOS. Some existing methods~\cite{perazzi2017learning, khoreva2017lucid, cheng2017segflow} model the \emph{short-term} consistency of the object appearance across neighboring frames in the video (Figure~\ref{fig:ls} (a)). The predicted mask of the previous frame is propagated to the current frame either by feature map aggregation or by optical-flow-guided pixel matching. The major issue with these methods is that they ignore the fine-grained interactions between local regions over time. In other words, the spatial regions in the predicted mask of the previous frames are integrated into the corresponding spatial regions in the current frame \emph{individually} without exploring their spatial-temporal correlations. As a result, local prediction errors may easily be propagated and amplified during temporal modeling, especially at the border of object regions. Ideally, we need a mechanism to model the fine-grained spatial-temporal interactions on the local frame regions, so that the consistency of object can be preserved.

Other works ~\cite{ventura2019rvos, xu2018youtube_s2s, tokmakov2017learning} apply \emph{Convolutional Recurrent Unit} to capture the evolution of the frame's convolutional feature map over a \emph{long-term time range}, and map the output of the recurrent units into a segmentation map of the current frame (Figure~\ref{fig:ls} (b)). These methods can get a long view of the video sequence preceding the current frame so that the long-range dynamics of the object can be captured, making the network be robust against occlusions and drift errors. Nevertheless, the major issue with these methods is that the feature maps fed into the recurrent units describe the holistic frame, which not only unnecessarily involves the background region into the learning process, but also dramatically increases the computational complexity. In fact, only the object mask regions are needed to model the evolution of object in time.

Motivated by the above issues in VOS, we propose an end-to-end \emph{Dual Temporal Memory Network} (DTMNet) that stores both short-term and long-term video sequence information as memories to assist the segmentation of a current frame (Figure.~\ref{fig:ls} (c)). In our network, the \emph{short-term memory sub-network} is designed as a spatial-temporal feature correlation module to capture the fine-grained inter-frame object appearance consistency. Given a current frame, we collect a small window of the preceding frames as its short-term memory. The frame and its memory frames are respectively encoded into a feature map in which each spatial location denotes one local region in the frame and the same feature location across different frames naturally encodes the evolution of a region across time. Then a spatial-temporal graph is built over all local regions in which each region is a node and the edges are established between regions within a local spatial-temporal window. The \emph{Graph Convolution}~\cite{jiang2019semi} operation is performed to update each region feature on the node according to its relations to others. By doing this, we model the spatial-temporal consistency of local regions across frames, leading to an improved segmentation performance.

The \emph{long-term sub-network} models the evolution of object across a long-time range. Given a current frame, we collect all preceding frames from the beginning of the video as its long-term memory. Instead of using the convolutional features of frames in the memory to model the dynamics of object over time, we propose to pool an object-orientated feature vector from the object mask on each frame, and apply the \emph{Simplified-Gated Recurrent Unit} (S-GRU) to learn a hidden-state vector to characterize the evolution of the object over a long-time range in the memory. This relieves the distractions of the background regions and significantly reduces the computational complexity.

The outputs from the short-term and the long-term sub-networks are sent to the \emph{segmentation sub-network} as supportive information to perform object segmentation. Extensive evaluations on three benchmark VOS datasets demonstrate that our DTMNet yields state-of-the-art performance in terms of both speed and accuracy. Our main contributions include:

(1) DTMNet for VOS, through which both the short-term spatial-temporal local region consistency and the long-term object evolution can be exploited.

(2) A graph-based learning framework to model the short-term spatial-temporal interactions of the local regions from neighboring frames in the video.

(3) An object-orientated feature based S-GRU module to model object evolution over a long-time range.

\begin{figure*}[tb]
\begin{center}
\begin{tabular}{c}
\includegraphics[width=0.89\linewidth]{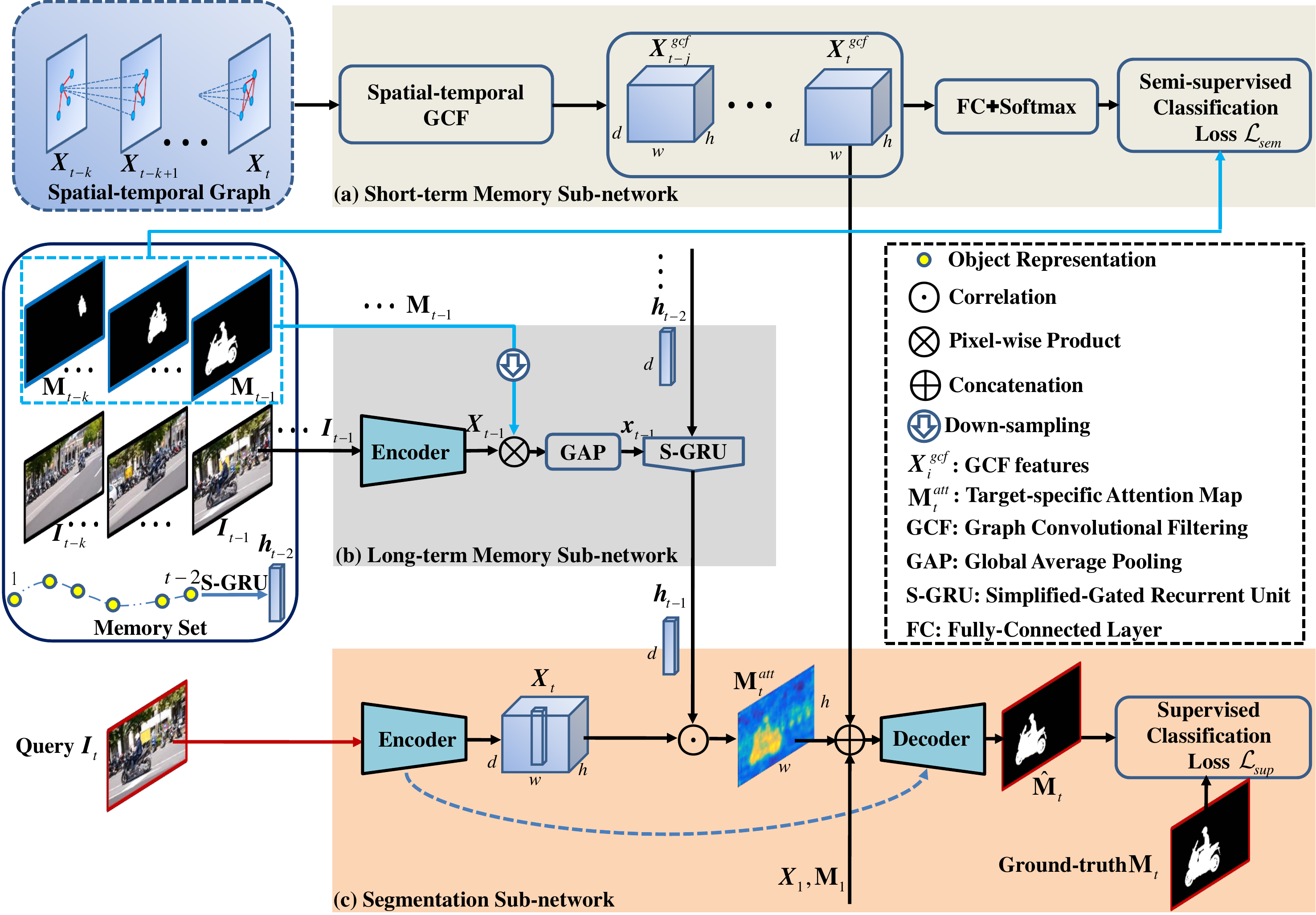}
\end{tabular}
\end{center}
   \caption{Pipeline of the proposed DTMNet for VOS. The network includes three key components: (a) A short-term memory sub-network to capture the spatial-temporal consistency of local regions over time; (b) A long-term memory sub-network to model the evolution of object over a long-time range to ensure robustness against occlusions and drift errors; (c) A segmentation sub-network that seamlessly fuses the short- and long-term memory information and the ground-truth information from the first frame to accurately predict the segmentation mask.}
\label{fig:Pipeline}
\end{figure*}
%
\section{Related Work}
\textbf{Video Object Segmentation}. There is a line of research on unsupervised VOS which leverages visual saliency~\cite{hu2018unsupervised,jang2016primary}, point trajectory~\cite{chen2015video} and motion~\cite{papazoglou2013fast} to segment objects from the background. Many semi-supervised VOS methods heavily rely on online fine-tuning on the first-frame mask to predict the masks on other frames during testing.
OSVOS~\cite{Caelles2017one-shot} and its extensions~\cite{voigtlaender2017online} ignore the temporal dimension and fine-tune a pre-trained fully convolutional network on the first frame to remember the object appearance. 
MHP-VOS~\cite{xu2019mhp} proposes a novel method called Multiple Hypotheses Propagation to defer the decision until a global view can be established.
Other methods take \emph{temporal information} into consideration. MSK~\cite{perazzi2017learning} and LucidTracker~\cite{khoreva2017lucid} use the predicted mask of the last frame as additional input of the current frame.
PReMVOS~\cite{luiten2018premvos} combines four different sub-networks to achieve impressive performance.
Although online fine-tuning boosts test accuracy, it badly sacrifices running efficiency for practical applications.

A growing line of research attempts to avoid the time-consuming online fine-tuning at the expense of a little accuracy reduction.
VideoMatch~\cite{hu2018videomatch} explores pixel-level embedding matching.
OSMN~\cite{yang2018efficient} uses two novel modulators to capture visual and spatial information of the target object and injects them into the segmentation branch.
FAVOS~\cite{cheng2018fast} utilizes tracking to obtain object bounding boxes and performs segmentation within the boxes.
AGAM-VOS~\cite{johnander2019generative} learns a probabilistic generative model to find a representation of the target and background appearance.
Our method shares the same spirit of not performing online fine-tuning as these methods, but the key difference is that we design a dedicated dual temporal memory mechanism to make the VOS accuracy even higher than some state-of-the-art online fine-tuning methods (see Table~\ref{table1}).

\textbf{Temporal Modeling in VOS}. Temporal sequence modeling plays an important role in VOS. Some methods try to model the \emph{long-term} object dynamics in the video sequence using Recurrent Neural Networks (RNNs).
RNN is designed to sequence modeling by propagating and accumulating a hidden state over time ~\cite{elman1990finding, kawakami2008supervised}. Gated Recurrent Unit (GRU) ~\cite{cho2014learning} and Long Short-Term Memory (LSTM) ~\cite{hochreiter1997long} are the two classic RNN components, both of which have been extended to CNNs. The existing methods introduce ConvLSTM and ConvGRU to model the long-term temporal dynamics in the video sequence. ~\cite{tokmakov2017learning} designs a visual memory module based ConvGRU to capture the evolution of object mask over times.
RVOS ~\cite{ventura2019rvos} presents a spatial-temporal recurrence module and applies ConvLSTM as the decoder. ~\cite{xu2018youtube_s2s} proposes a sequence-to-sequence network by ConvLSTM.
Without any RNN unit, STCNN~\cite{xu2019spatiotemporal} designs a novel temporal coherence branch inspired by video predict task, and is also able to model the long-term dynamics in video. STMN~\cite{oh2019video} stores the first, intermediate and previous frame in the memory and uses them as reference to infer the object mask of the current frame.  The advantage of modeling long-term temporal dynamics is to allow the network to get the long view of the video sequence preceding the current frame, making the network be robust against occlusions and drift errors.

The other VOS methods model the short-term visual consistency across neightboring frames in the video. MSK~\cite{perazzi2017learning} uses the predicted mask of the last frame as additional input of the current frame to help with the mask prediction. RGMP~\cite{wug2018fast} utilizes a Siamese encoder-decoder network to extract the first, previous and current feature to propagate the previous predicted mask to current frame. Optical flow is also commonly used to match the pixel correspondence between successive frames, through which the object mask of sequential frames can be estimated~\cite{khoreva2017lucid, cheng2017segflow, luiten2018premvos}. It turns out that modeling short-term visual consistency is an important prior to enhance the performance of VOS, and is commonly applied in the VOS works~\cite{perazzi2017learning,wug2018fast,khoreva2017lucid, cheng2017segflow, luiten2018premvos}. In contrast to these existing methods, we integrate the long- and short-term temporal modeling into a unified framework, and each sub-network in our network design is dedicated to resolve the issues of the existing methods.
\section{Proposed Method}
%
%
%
\subsection{Framework Overview}
\label{sec:overview}
Given a video sequence with $t$ frames $\mathcal{I}_1^t=\{\textit{\textbf{I}}_i\}_{i=1}^t$ and the binary ground-truth mask $\textbf{M}_1\in\{0,1\}^{w\times h}$ of the first frame $\textit{\textbf{I}}_1$ with width $w$ and height $h$, our task is to predict the segmentation masks of the subsequent frames $\mathcal{I}_2^t$, denoted as $\mathcal{M}_2^t=\{\textbf{M}_i\}_{i=2}^t$.
To this end, we develop the DTMNet for VOS, as illustrated by Figure~\ref{fig:Pipeline}.
Our DTMNet is composed of three seamless components: (a) A short-term memory sub-network; (b) A long-term memory sub-network and (c) A segmentation sub-network.
Among them, the segmentation module takes full advantages of the complementary characteristics of the rich supportive information provided by the short- and the long-term memory modules, \textit{i.e.,} good adaptation to appearance changes and robustness against occlusions and drift errors, thereby enabling to predict an accurate segmentation mask.
Specifically, when segmenting video frame $\textit{\textbf{I}}_t$, we take it as the query frame and the preceding $k$ frames $\mathcal{I}_{t-k}^{t-1}$ with their masks $\mathcal{M}_{t-k}^{t-1}$ as the short-term memories.
The frames $\mathcal{I}_{t-k}^{t}$ are fed into the backbone network as the encoder to extract features $\mathcal{X}_{t-k}^{t}=\{\textit{\textbf{X}}_i\}_{i=t-k}^t$, where the feature map $\textit{\textbf{X}}_i\in \mathbb{R}^{w\times h\times d}$ with $d$ channels.
Afterwards, as shown in Figure~\ref{fig:Pipeline}(a), the features $\mathcal{X}_{t-k}^{t}$ are fed into a spatial-temporal graph convolutional filtering (GCF) module, generating the refined features $\textit{\textbf{X}}_t^{gcf}\in \mathbb{R}^{w\times h\times d}$ for the query frame $\textit{\textbf{I}}_t$.
The GCF leverages Laplacian smoothing to compute $\textit{\textbf{X}}_t^{gcf}$ that can be viewed as a low-pass filtering process~\cite{li2018deeper}.
The smoothing makes the features in the same cluster similar, facilitating the subsequent classification task in the segmentation sub-network.
Meanwhile, as shown by Figure~\ref{fig:Pipeline}(b), we leverage S-GRU to model the long-term memory that simplifies the GRU proposed by~\cite{cho2014learning} with only one update gate left.
The output of S-GRU is a $d$-dimensional hidden-state vector $\textit{\textbf{h}}_{t-1}\in \mathbb{R}^d$ that can memorize all object appearances $\{\textit{\textbf{x}}_i\}_{i=1}^{t-1}$ appearing before frame $\textit{\textbf{I}}_t$, where $\textit{\textbf{x}}_i\in \mathbb{R}^d$ denotes the object representation at frame $\textit{\textbf{I}}_i$.
The S-GRU updates its states $\textit{\textbf{h}}_1\stackrel{\textit{\textbf{x}}_2}{\longrightarrow} \textit{\textbf{h}}_2\cdots\stackrel{\textit{\textbf{x}}_{t-1}}{\longrightarrow}\textit{\textbf{h}}_{t-1}$ incrementally across the video frames, which can effectively capture the long-range dynamics of the object that are robust against occlusions and drifting with less memory overhead.

Finally, as shown in Figure~\ref{fig:Pipeline}(c), the learned hidden-state vector $\textbf{\textit{h}}_{t-1}$ is correlated with the query image features $\textit{\textbf{X}}_t$ to generate a target-specific attention map $\textbf{M}_t^{att}\in \mathbb{R}^{w\times h}$, which highlights the target-specific region while suppressing other distractors. Then, we concatenate the attention map $\textbf{M}_t^{att}$ and the GCF features $\textit{\textbf{X}}_t^{gcf}$ to further refine the target-specific features of the query image.
Moreover, we also concatenate features $\textit{\textbf{X}}_1$ and its ground-truth mask $\textbf{M}_1$ from the first frame to further strengthen the target-specific feature representation. Finally, the concatenated features $\textbf{M}_t^{att}\oplus\textit{\textbf{X}}_t^{gcf}\oplus\textbf{M}_1\oplus\textit{\textbf{X}}_1$ are fed into the decoder module to produce the final segmentation mask $\hat{\textbf{M}}_t$, with skip-connections to fuse multi-scale features of different layers like U-Net~\cite{ronneberger2015u}.
\subsection{Short-term Memory Sub-network}
\label{sec:short-term}
As aforementioned, the short-term memory sub-network is to model the inter-frame temporal correlation.
Previous works achieve this by propagating the mask from previous frame to current frame, either by directly concatenating the previously predicted mask and the current frame~\cite{perazzi2017learning,wug2018fast,johnander2019generative,voigtlaender2019feelvos} or depending on the optical-flow guided pixel matching between two sequential frames~\cite{hu2017maskrnn,cheng2017segflow,tokmakov2017learning,linagss}.
However, the former is easy to introduce noisy backgrounds into the object regions, especially on the object boundaries, leading to sub-optimal accuracy.
Although the latter seems reasonable, there exist two limitations: First, it is very computationally expensive to estimate optical flows.
Second, estimating optical flows needs to compute point-to-point mapping between two pixels, which is too restrictive~\cite{li2018low}.
For the high-level feature maps, they involve both the strength of the responses and their spatial locations~\cite{he2015spatial}, where each feature corresponds to a single site in the predicted mask. Hence, mask propagation can be implemented via feature propagation.
Moreover, due to the fact that each feature in the high-level feature maps represents a local region inside the receptive field of the CNN filter instead of a single image pixel, a linear combination of these features to implement feature propagation serves well to model the spatial-temporal interactions between the local regions across video frames, thereby enabling to well preserve their spatial-temporal consistency across the frames.
Motivated by this analysis, we propose to propagate features by spatial-temporal GCF, that is, using graph convolutions to linearly combine spatial-temporal neighbors.

\textbf{Notations.} As shown in Figure~\ref{fig:Pipeline}(a), given the short-term memory set $\mathcal{X}_{t-k}^{t}$, we define the spatial-temporal graph as $\mathcal{G}=(\mathcal{V},\mathcal{E},\textbf{{A}},\textbf{X})$~\cite{wu2019comprehensive}, where $\mathcal{V}=\{v_i\}_{i=1}^N$ denotes the node set with size $N=(k+1)wh$, $\mathcal{E}=\{e_{ij}\}_{i,j=1}^N$ is the edge set, where $e_{ij}$ models the pairwise relations between any two nodes $i$ and $j$, $\textbf{{A}}\in \mathbb{R}^{N\times N}$ is the adjacency matrix whose entry $\textbf{{A}}(i,j)$ is the weight of edge $e_{ij}$,
$\textbf{X}=[\textit{\textbf{x}}_1^\top;\ldots;\textit{\textbf{x}}_N^\top]\in \mathbb{R}^{N\times d}$ is the feature matrix constructed by set $\mathcal{X}_{t-k}^{t}$, where $\textit{\textbf{x}}_i\in \mathbb{R}^d$ denotes the feature representation of node $v_i$.

\textbf{Sparse Adjacency Matrix ${\textbf{A}}$.} For each node $v_i\in \mathcal{V}$, we construct its edge set $\{e_{ij},j\in \mathcal{N}(i)=\mathcal{N}^s(i)\cup \mathcal{N}^t(i)\}$ to capture the spatial-temporal interactions of the pair-wise local regions across video frames,
where $\mathcal{N}^s(i)$ is an $w_s\times h_s$ window centered at node $v_i$ at the current frame,
$\mathcal{N}^t(i)$ denotes an $w_t\times h_t$ window centered at node $v_i$ at the next frame.
The number of none-zero edges in $\mathcal{E}$ is $N(w_sh_s+w_th_t)\ll N^2$, leading to a sparse ${\textbf{A}}$ with less computational cost.
To learn task-specific similarity between nodes $i$ and $j$ for adaptive graph learning, we define the weight of edge $e_{ij}$ as
\begin{equation}
\label{eq:Aij}
\textbf{A}(i,j) =\sigma((\textbf{W}_1\textit{\textbf{x}}_i)^\top(\textbf{W}_2\textit{\textbf{x}}_j)),
\end{equation}
where $\sigma(x)=\frac{1}{1+e^{-x}}, x\in \mathbb{R}$ is the sigmoid function, $\textit{\textbf{x}}_i, \textit{\textbf{x}}_j\in \mathbb{R}^d$ denote node features, $\textbf{W}_1,\textbf{W}_2\in \mathbb{R}^{r\times d}$ are learnable weight matrices.
\begin{figure}
\begin{center}
\begin{tabular}{c}
\includegraphics[width=0.78\linewidth]{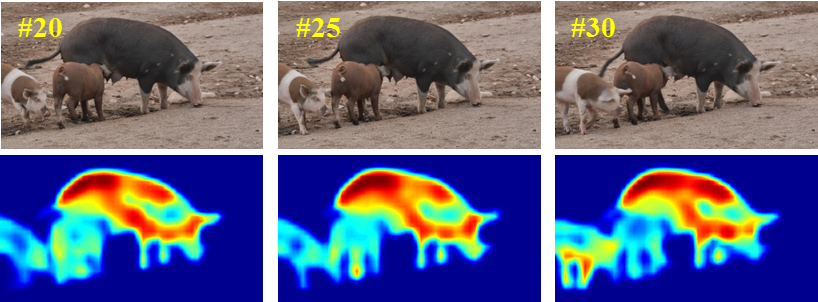}
\end{tabular}
\end{center}
   \caption{Laplacian smoothing effect of GCF features. Top: three query frames selected from sequence \textit{pigs} in DAVIS 2017 val dataset~\cite{Pont-Tuset_arXiv_2017}; bottom: the corresponding GCF feature responses, showing that the features of the same object across frames well preserve spatial-temporal consistency.}
\label{fig:short-term}
\end{figure}

\textbf{Spatial-temporal GCF.} To perform GCF, we apply the graph convolutional networks (GCNs) proposed in~\cite{kipf2016semi}.
We design one-layer graph convolution in our DTMNet as
\begin{equation}
\label{eq:layer-graph}
 \textit{\textbf{z}} = \textbf{X}^{gcf}\textit{\textbf{w}},
\end{equation}
where $\textit{\textbf{w}}\in \mathbb{R}^{d}$ denotes the weight vector of the FC layer and
%
\begin{equation}
\label{eq:laplaciansmoothing}
\textbf{X}^{gcf} = \tilde{\textbf{D}}^{-\frac{1}{2}}\tilde{\textbf{A}}\tilde{\textbf{D}}^{-\frac{1}{2}}\textbf{X},
\end{equation}
where $\tilde{\textbf{A}}={\textbf{A}}+\textbf{I}$, $\textbf{I}$ denotes the identity matrix, $\tilde{\textbf{D}}$ is a diagonal matrix with $\tilde{\textbf{D}}(i,i)=\sum_j{\tilde{\textbf{A}}(i,j)}$.

After GCF using (\ref{eq:laplaciansmoothing}), the $i$-th node representation $\textit{\textbf{x}}_i^{gcf}$ can be formulated as
\begin{equation}
\label{eq:xgcf}
\textit{\textbf{x}}_i^{gcf} = \sum_{j=1}^{|\mathcal{N}(i)|}\frac{\textbf{A}(i,j)}{\sqrt{\tilde{\textbf{D}}(i,i)\tilde{\textbf{D}}(j,j)}}\textit{\textbf{x}}_j+\textit{\textbf{x}}_i.
\end{equation}
It is obvious that $\textit{\textbf{x}}_i^{gcf}$ in (\ref{eq:xgcf}) is a linear combination of the nodes in its spatial-temporal neighborhood $\mathcal{N}(i)$, thereby expressing
feature propagation more accurately than existing optical-flow-based methods~\cite{hu2017maskrnn,cheng2017segflow,tokmakov2017learning,linagss} that are limited by restrictive point-to-point mappings.
Moreover, (\ref{eq:xgcf}) is a Laplacian smoothing process~\cite{li2018deeper} that calculates the new features $\textit{\textbf{x}}_i^{gcf}$ as the weighted average of its neighboring features in $\mathcal{N}(i)$ and itself $\textit{\textbf{x}}_i$.
The smoothing makes the features in the same cluster similar that favorably preserves the spatial-temporal consistency of the segmented object across frames as illustrated by Figure~\ref{fig:short-term}, rendering a great benefit to the downstream pixel-wise classification task in the segmentation sub-network (\S~\ref{sec:segmentation}).
%

\textbf{Semi-supervised Classification.}
When training our model, we assume that in set $\mathcal{X}_{t-k}^t$,
the ground-truth masks $\mathcal{M}_{t-k}^{t-1}$ are given that correspond to the labeled nodes $\mathcal{V}_l\in \mathcal{V}$, while the query image mask $\textbf{M}_t$ is to be propagated from nodes $\mathcal{V}_l$.
We leverage a one-layer GCN which applies a softmax classifier on the output features $\textit{\textbf{z}}$ in (\ref{eq:layer-graph})
\begin{equation}
\hat{\textit{\textbf{y}}} = \mathrm{softmax}(\textit{\textbf{z}}).
\end{equation}
The loss function is defined as the cross-entropy error over all the labeled nodes
\begin{equation}
\label{eq:semi-loss}
\mathcal{L}_{sem} = -\sum_{i\in \mathcal{V}_l}\textit{\textbf{y}}(i)\log(\hat{\textit{\textbf{y}}}(i)),
\end{equation}
where $\textit{\textbf{y}}(i)\in \{0,1\}$ denotes the ground-truth label of node $i\in\mathcal{V}_l$.
%
%
\subsection{Long-term Memory Sub-network}
\label{sec:long-term}
Using short-term memory for VOS can deal with target appearance changes well. However, it suffers from drifting problem under challenging scenarios such as severe occlusion or fast motion between sequential frames.
To address this issue, we further develop the S-GRU module to capture long-term memory information as a complement, as illustrated by Figure~\ref{fig:Pipeline}(b).

For frame $\textit{\textbf{I}}_{t-1}$, given its features $\textit{\textbf{X}}_{t-1}$ and segmentation mask $\textbf{M}_{t-1}$, we first mask out object features as $\textit{\textbf{X}}_{t-1}\otimes \textbf{M}_{t-1}$, where $\otimes$ denotes pixel-wise product.
Then, we feed the object features into a global average pooling (GAP) layer, yielding
\begin{equation}
\label{eq:input-GUR}
\textit{\textbf{x}}_{t-1} = \mathrm{GAP}(\textit{\textbf{X}}_{t-1} \otimes \textbf{M}_{t-1}),
\end{equation}
which captures global context information that is robust against object appearance variations.
Next, the S-GRU leverages $\textit{\textbf{x}}_{t-1}$ in (\ref{eq:input-GUR}) and the previous state $\textit{\textbf{h}}_{t-2}$ to compute the new state $\textit{\textbf{h}}_{t-1}$.
The state vector $\textit{\textbf{h}}$ plays a key role in S-GRU since it well captures the long-term dynamics of the object across frames.
Then, the learning process is formulated as
\begin{equation}
\label{eq:ourGRU}
\begin{aligned}
&\textit{\textbf{z}}_{t-1} = \sigma(\textbf{W} [\textit{\textbf{x}}_{t-1};\textit{\textbf{h}}_{t-2}]),\\
&\textit{\textbf{h}}_{t-1} = (1-\textit{\textbf{z}}_{t-1})\odot \textit{\textbf{h}}_{t-2} + \textit{\textbf{z}}_{t-1}\odot \textit{\textbf{x}}_{t-1},
\end{aligned}
\end{equation}
where $\odot$ denotes the element-wise multiplication, $\sigma$ is the sigmoid function, $\textbf{W}\in \mathbb{R}^{d\times 2d}$ is a learnable weight matrix.
Different from the ConvGRU~\cite{tokmakov2017learning} that consists of update and reset gates, our S-GRU in (\ref{eq:ourGRU}) only has update gate $\textit{\textbf{z}}_{t-1}$, which reduces computational complexity significantly.
In (\ref{eq:ourGRU}), the new state $\textit{\textbf{h}}_{t-1}$ is a weighted sum of the current object representation $\textit{\textbf{x}}_{t-1}$ and the previous state $\textit{\textbf{h}}_{t-2}$ that memorizes the dynamic object appearances across all previous frames. If the update gate $\textit{\textbf{z}}_{t-1}$ is close to one, the memories encoded in $\textit{\textbf{h}}_{t-2}$ will be forgotten.
\begin{table*}[t]
\footnotesize
\caption{Comparison of our DTMNet with the state of the arts on DAVIS 2016 val. \textcolor[rgb]{1.00,0.00,0.00}{\textbf{Red}} and \textcolor[rgb]{0.00,0.00,1.00}{{\textbf{blue}}} bold fonts indicate the best, the second-best performance respectively. }
\setlength{\tabcolsep}{1.4mm}
\begin{center}
\begin{tabular}{|c||c|c|ccc|ccc|c|}
\hline

Method   &OL &$\mathcal{J\&F}$ $\uparrow$  &$\mathcal{J}$ Mean $\uparrow$ &$\mathcal{J}$ Recall $\uparrow$ &$\mathcal{J}$ Decay $\downarrow$ &$\mathcal{F}$ Mean $\uparrow$ &$\mathcal{F}$ Recall $\uparrow$ &$\mathcal{F}$ Decay $\downarrow$ &Time (s) $\downarrow$ \\
\hline
MSK~\cite{perazzi2017learning}   &\Checkmark &77.6  &79.7 &93.1 &8.9 &75.4 &87.1 &9.0 &12  \\
LIP~\cite{lyu2019lip}   &\Checkmark &78.5  &78.0 &88.6 &\color{blue}\textbf{5.0} &79.0 &86.8 &\color{blue}\textbf{6.0} &-  \\
OSVOS~\cite{Caelles2017one-shot} &\Checkmark &80.2  &79.8 &93.6 &14.9 &80.6 &{92.6} &15.0 &9  \\
Lucid~\cite{khoreva2017lucid}   &\Checkmark &83.6  &84.8 &- &- &82.3 &- &- &\textgreater 30  \\
STCNN~\cite{xu2019spatiotemporal}   &\Checkmark &83.8  &83.8 &{96.1} &\color{red}\textbf{4.9} &83.8 &91.5 &{6.4} &3.9  \\
CINM~\cite{bao2018cnn}   &\Checkmark &84.2  &83.4 &94.9 &12.3 &85.0 &92.1 &14.7 &\textgreater 30  \\
OnAVOS~\cite{voigtlaender2017online} &\Checkmark &85.5  &\color{red}\textbf{86.1} &{96.1} &{5.2} &84.9 &89.7 &\color{red}\textbf{5.8} &13  \\
OSVOS-S~\cite{maninis2018video}   &\Checkmark &{86.6}  &{85.6} &\color{red}\textbf{96.8} &5.5 &{87.5} &\color{red}\textbf{95.9} &8.2 &4.5  \\
PReMVOS~\cite{luiten2018premvos}   &\Checkmark &\color{blue}\textbf{86.8}  &84.9 &{96.1} &8.8 &\color{red}\textbf{88.6} &94.7 &9.8 &\textgreater 30  \\
MHP-VOS~\cite{xu2019mhp} &\Checkmark &\color{red}\textbf{86.9}  &\color{blue}\textbf{85.7} &\color{blue}\textbf{96.6} &- &\color{blue}\textbf{88.1} &\color{blue}\textbf{94.8} &- &$>$14  \\
\hline
\hline
VPN~\cite{jampani2017video}   &\XSolidBrush &67.9  &70.2 &82.3 &12.4 &65.5 &69.0 &14.4 &0.63  \\
OSMN~\cite{yang2018efficient}   &\XSolidBrush &73.5  &74.0 &87.6 &{9.0} &72.9 &84.0 &10.6 &0.14  \\
VideoMatch~\cite{hu2018videomatch}   &\XSolidBrush &-  &81.0 &- &- &- &- &- &0.32  \\
FAVOS~\cite{cheng2018fast}   &\XSolidBrush &81.0  &\color{blue}\textbf{82.4} &\color{red}\textbf{96.5} &\color{red}\textbf{4.5} &79.5 &89.4 &\color{red}\textbf{5.5} &1.8  \\
FEELVOS~\cite{voigtlaender2019feelvos}   &\XSolidBrush &{81.7}  &81.1 &90.5 &13.7 &\color{blue}\textbf{82.2} &86.6 &14.1 &0.45\\
RGMP~\cite{wug2018fast}   &\XSolidBrush &\color{blue}\textbf{81.8}  &{81.5} &91.7 &10.9 &82.0 &\color{blue}\textbf{90.8} &10.1 &{0.13}  \\
AGAM-VOS~\cite{johnander2019generative}   &\XSolidBrush &\color{blue}\textbf{81.8}  &81.4 &{93.6} &9.4 &82.1 &{90.2} &{9.8} &\color{red}\textbf{0.07}  \\
\textbf{DTMNet}   &\XSolidBrush &\color{red}\textbf{85.4}  &\color{red}\textbf{85.9} &\color{blue}\textbf{96.0} &\color{blue}\textbf{4.7} &\color{red}\textbf{84.9} &\color{red}\textbf{92.0} &\color{blue}\textbf{5.7} &\color{blue}\textbf{0.12}  \\
\hline
\end{tabular}
\end{center}
\label{table1}
\end{table*}
\begin{figure}
\begin{center}
\begin{tabular}{c}
\includegraphics[width=0.8\linewidth]{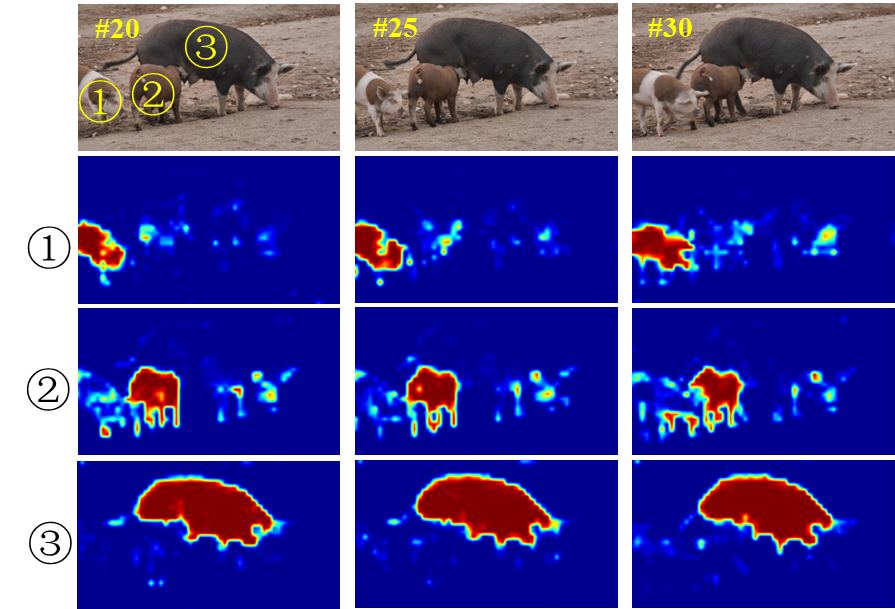}
\end{tabular}
\end{center}
   \caption{Illustration of the target-specific attention maps. The same query frames are shown in Figure~\ref{fig:short-term}, including three pigs (ID numbers: \textcircled{1}, \textcircled{2}, \textcircled{3}) to be segmented.}
\label{fig:target-attention}
\end{figure}
\subsection{Segmentation Sub-network}
\label{sec:segmentation}
Figure~\ref{fig:Pipeline}(c) illustrates the architecture of our segmentation sub-network. Similar to U-Net~\cite{ronneberger2015u}, our segmentation network uses skip-connections to fuse multi-scale features from the encoder to the decoder modules.
The encoder uses the ResNet101 backbone network~\cite{he2016deep} with dilated convolutions to set the stride of the deepest layer to 16.
For query frame $\textit{\textbf{I}}_{t}$, the deepest layer outputs feature maps $\textit{\textbf{X}}_{t}$. Then, we correlate $\textit{\textbf{X}}_{t}$ with the hidden-state vector $\textit{\textbf{h}}_{t-1}$ learned from the long-term memory sub-network, yielding the target-specific attention map $\textbf{M}_t^{att}=\textit{\textbf{X}}_{t}\odot \textit{\textbf{h}}_{t-1}$.
As illustrated by Figure~\ref{fig:target-attention}, the learned $\textbf{M}_t^{att}$ can effectively highlight the target-specific regions while suppressing other distractors such as other objects and backgrounds.
Next, after achieving the GCF features $\textit{\textbf{X}}_t^{gcf}$ from the short-term memory sub-network, we feed the concatenated features $\textit{\textbf{X}}_t^{gcf}\oplus \textbf{M}_t^{att}\oplus \textit{\textbf{X}}_1\oplus  \textbf{M}_1$ into the decoder.
Meanwhile, we leverage skip-connections to concatenate the feature maps from the decoder and its counterparts from the encoder, which are then gradually upscaled by a factor of two at a time. Afterwards, they are concatenated with the following layer features.
Finally, the aggregated features are fed into a convolutional layer following a softmax layer to predict the object mask $\hat{\textbf{M}}_t$.
As the short-term memory sub-network, the loss function here is also defined as the cross-entropy loss for pixel-wise classification task:
\begin{equation}
\label{eq:sup-loss}
\mathcal{L}_{sup} = -\sum_{t}\sum_{ij} \textbf{M}_t(i,j)\log \hat{\textbf{M}}_t(i,j),
\end{equation}
where $\textbf{M}_t\in\{0,1\}^{w\times h}$ denotes the ground-truth mask of frame $\textit{\textbf{I}}_t$.

Finally, the loss function for the whole network training is defined as
\begin{equation}
\label{eq:loss}
\mathcal{L} = \mathcal{L}_{sem}+\lambda\mathcal{L}_{sup},
\end{equation}
where $\mathcal{L}_{sem}$ is defined in (\ref{eq:semi-loss}), $\lambda>0$ is a pre-defined trade-off parameter.
\section{Experimental Results}

%
\subsection{Implementation Details}
\label{sec:implementation}
Following AGAME-VOS~\cite{johnander2019generative}, the training process of our DTMNet is divided into two stages:

\textbf{Stage 1.} Firstly, we train our DTMNet using the Adam optimizer~\cite{kingma2014adam} to minimize the loss $\mathcal{L}$ in (\ref{eq:loss}) on DAVIS 2017~\cite{Pont-Tuset_arXiv_2017} and YouTube-VOS~\cite{xu2018youtube} datasets for $80$ epochs, where all training images are resized to $240\times432$ pixels.
Each batch contains $4$ videos, where $8$ frames are randomly selected for training in each video.
The hyperparameters in our DTMNet are set empirically as learning rate $= 1e-4$, learning rate decay $= 0.95$ and weight decay $= 1e-5$.

\textbf{Stage 2.} Next, we fine-tune the trained model at \textbf{Stage 1} on the same datasets for $100$ epochs but the images are resized to $480\times864$ pixels which is twice the size of the input images at \textbf{Stage 1}.
Each batch contains $2$ videos with randomly selected $5$ frames in each video. The parameters are also set empirically as learning rate $= 1e-5$, learning rate decay = $0.985$ and weight decay $= 1e-6$.

The DTMNet is implemented in Pytorch and an Nvidia GTX 2080Ti is used for acceleration. All of the training procedures can be completed within one day.
\subsection{Datasets and Evaluation Metrics}\label{sec:setup}

\textbf{Datasets.} We train and evaluate the DTMNet on three VOS benchmark datasets, including DAVIS 2016~\cite{Perazzi2016},
DAVIS 2017~\cite{Pont-Tuset_arXiv_2017} and YouTube-VOS~\cite{xu2018youtube}.
The DAVIS 2016 is a densely-annotated VOS dataset, which contains $30$ training and $20$ validation video sequences of high-quality with $3,455$ highly accurate pixel-wise annotation in total.
The DAVIS 2017 enlarges the DAVIS 2016 by introducing more additional videos with multi-objects.
The DAVIS 2017 contains a training set with $60$ sequences, a validation set with $30$ sequences, a test-dev set with $30$ sequences and a test-challenge set with $30$ sequences.
%
%
The YouTube-VOS is the first large-scale VOS dataset, which is more than $30$ times larger than existing largest dataset at that time.
The YouTube-VOS consists of $3,471$ videos in the training set, $474$ videos in the validation set with $65$ seen categories, and $26$ unseen categories in the training set.

\textbf{Evaluation Metrics.} We use the standard metrics provided by the DAVIS challenge~\cite{Pont-Tuset_arXiv_2017}, including the region similarity $\mathcal{J}$, contour accuracy $\mathcal{F}$ and the mean of the two metrics $\mathcal{J\&F}$.
Given the estimated segmentation mask $\hat{\textbf{M}}$ and the ground-truth mask $\textbf{M}$, the region similarity is calculated as $\mathcal{J}={\frac{\left|\hat{\textbf{M}}\cap \textbf{M}\right|}{\left|\hat{\textbf{M}}\cup \textbf{M}\right|}}$.
The contour accuracy is measured by the F-measure $\mathcal{F}$ between the contour-based precision $\mathcal{P}$ and recall $\mathcal{R}$ as  $\mathcal{F}=\frac{2\mathcal{PR}}{\mathcal{P+R}}$.
%
\subsection{Comparison with the State-of-the-arts}\label{sec:comparisions}
We compare our DTMNet with some state-of-the-art online-learning (OL) VOS methods and some offline ones on the DAVIS 2016, the DAVIS 2017 and the YouTube-VOS benchmark datasets.
It is worth noting that the our DTMNet does not resort to any post-processing or OL technique.

\textbf{Results on DAVIS 2016.}
Table~\ref{table1} lists the evaluation results on DAVIS 2016 by our DTMNet and $17$ state-of-the-art OL and offline VOS methods in comparison.
Among the offline methods, our DTMNet achieves the best performance in terms of $\mathcal{J\&F}$ ($85.4\%$), $\mathcal{J}$ Mean ($85.9\%$), $\mathcal{F}$ Mean ($84.9\%$) and $\mathcal{F}$ Recall ($92.0\%$), the second-best performance with $\mathcal{J}$ Recall of $96.0\%$, $\mathcal{J}$ Decay of $4.7\%$ and $\mathcal{F}$ Decay of $5.7\%$. Furthermore, the DTMNet is the first runner-up with a fast speed of $0.12$ s/frame, closely following the AGAM-VOS that runs at $0.07$ s/frame.
Even compared with the OL methods, the DTMNet still has a competing $\mathcal{J\&F}$ of $85.4\%$, which is only slightly lower than the best-performing MHP-VOS with $\mathcal{J\&F}$ of $86.9\%$  by $1.5\%$. Besides, the DTMNet runs at $0.12$ s/frame, which is much faster than the MHP-VOS at a speed of more than $14$ s/frame.
%

%
\begin{table}[t]
\footnotesize
\caption{Comparison  of our DTMNet with the state of the arts on DAVIS 2017 val.}
\setlength{\tabcolsep}{1.4mm}
\begin{center}
\begin{tabular}{|c||c|c|c|c|c|}
\hline

Method   &OL &$\mathcal{J\&F}$ $\uparrow$  &$\mathcal{J}$ Mean $\uparrow$ &$\mathcal{F}$ Mean $\uparrow$ &Time (s) $\downarrow$\\
\hline
MSK~\cite{perazzi2017learning} &\Checkmark &54.3  &51.2 &57.3 &15\\
OSVOS~\cite{Caelles2017one-shot}&\Checkmark &60.3  &56.6 &63.9 &\color{blue}\textbf{11}\\
LIP~\cite{lyu2019lip}   &\Checkmark &61.1  &59.0 &63.2 &-\\
STCNN~\cite{xu2019spatiotemporal} &\Checkmark &61.7  &58.7 &64.6 &6\\
OnAVOS~\cite{voigtlaender2017online}   &\Checkmark &65.4  &61.6 &69.1 &26\\
OSVOS-S~\cite{maninis2018video}   &\Checkmark &68.0  &64.7 &71.3 &\color{red}\textbf{8}\\
CINM~\cite{bao2018cnn}   &\Checkmark &\color{blue}\textbf{70.6}  &\color{blue}\textbf{67.2} &\color{blue}\textbf{74.0} &50\\
MHP-VOS~\cite{xu2019mhp} &\Checkmark &\color{red}\textbf{75.3}  &\color{red}\textbf{71.8} &\color{red}\textbf{78.8} &20\\
\hline
\hline
OSMN~\cite{yang2018efficient}   &\XSolidBrush &54.8  &52.5 &57.1 &0.28\\
SiamMask~\cite{wang2019fast}   &\XSolidBrush &56.4 &54.3 &58.5 &0.02\\
FAVOS~\cite{cheng2018fast}   &\XSolidBrush &58.2  &54.6 &61.8 &1.2\\
VideoMatch~\cite{hu2018videomatch}   &\XSolidBrush &62.4  &56.6 &68.2 &0.35\\
RANet~\cite{wang2019ranet} &\XSolidBrush &65.7  &63.2 &68.2 &-\\
RGMP~\cite{wug2018fast}   &\XSolidBrush &66.7  &64.8 &68.6 &0.28\\
AGSS-VOS~\cite{linagss}   &\XSolidBrush &67.4  &64.9 &69.9 &-\\
AGAM-VOS~\cite{johnander2019generative}   &\XSolidBrush &70.0  &67.2 &72.7 &-\\
DMM-Net~\cite{zeng2019dmm}   &\XSolidBrush &{70.7}  &\color{blue}\textbf{68.1} &73.3 &\color{red}\textbf{0.13}\\
FEELVOS~\cite{voigtlaender2019feelvos}   &\XSolidBrush &\color{red}\textbf{71.6}  &\color{red}\textbf{69.1} &\color{red}\textbf{74.0} &0.51\\
\textbf{DTMNet}   &\XSolidBrush &\color{blue}\textbf{71.5}  &\color{red}\textbf{69.1} &\color{blue}\textbf{73.9} &\color{blue}\textbf{0.17}\\
\hline
\end{tabular}
\end{center}
\label{table2}
\end{table}
\textbf{Results on DAVIS 2017.}
The DAVIS 2017 considers multi-object scenarios, making it more challenging than the DAVIS 2016 that is only for single-object segmentation.
Table~\ref{table2} lists the comparison results of our DTMNet with $18$ state-of-the-art OL and off-line methods.
%
Among them, we can observe that our DTMNet has the best performance in terms of $\mathcal{J}$ Mean ($69.1\%$), and the second-best $\mathcal{J}\&\mathcal{F}$ of $71.5\%$ and $\mathcal{F}$ Mean of $73.9\%$, closely following the best-performing FEELVOS in terms of $\mathcal{J}\&\mathcal{F}$ ($71.6\%$) and $\mathcal{F}$ Mean ($74.0\%$) with only a small gap of $0.1\%$, but our DTMNet runs at $0.17$ s/frame on DAVIS 2017 val, which is much faster than FEELVOS that is $0.51$ s/frame.
Furthermore, the DTMNet even outperforms the second best-performing offline method CINM in terms of $\mathcal{J}\&\mathcal{F}$ and $\mathcal{J}$ Mean by $0.9\%$ and $1.9\%$, respectively, demonstrating the effectiveness of the dual temporal memory learning strategy in our DTMNet.
\begin{table}[t]
\footnotesize
\caption{Comparison  of our DTMNet with the state of the arts on YouTube-VOS dataset.}
\setlength{\tabcolsep}{1.4mm}
\begin{center}
\begin{tabular}{|c||c|c|c|c|c|c|}
\hline
Method   &OL &$\mathcal{G}\uparrow$  &$\mathcal{J}_{s}\uparrow$ &$\mathcal{F}_{s}\uparrow$ &$\mathcal{J}_{u}\uparrow$ &$\mathcal{F}_{u}\uparrow$\\
\hline
MSK~\cite{perazzi2017learning}    &\Checkmark &53.1  &59.9 &59.5 &45.0 &47.9\\
OnAVOS~\cite{voigtlaender2017online}   &\Checkmark &55.2  &\color{blue}\textbf{60.1} &\color{blue}\textbf{62.7} &46.6 &51.4\\
OSVOS~\cite{Caelles2017one-shot}   &\Checkmark &\color{blue}\textbf{58.8}  &59.8 &60.5 &\color{blue}\textbf{54.2} &\color{blue}\textbf{60.7}\\
S2S~\cite{xu2018youtube_s2s}   &\Checkmark &\color{red}\textbf{64.4}  &\color{red}\textbf{71.0} &\color{red}\textbf{70.0} &\color{red}\textbf{55.5} &\color{red}\textbf{61.2}\\
\hline
\hline
OSMN~\cite{yang2018efficient}   &\XSolidBrush &51.2  &60.0 &60.1 &40.6 &44.0\\
DMM-Net~\cite{zeng2019dmm}   &\XSolidBrush &51.7  &58.3 &60.7 &41.6 &46.3\\
SiamMask~\cite{wang2019fast}   &\XSolidBrush &52.8 &60.2 &58.2 &45.1 &47.7\\
RGMP~\cite{wug2018fast}   &\XSolidBrush &53.8  &59.5 &- &45.2  &-\\
RVOS~\cite{ventura2019rvos}   &\XSolidBrush &56.8  &63.6 &67.2 &45.5 &51.0\\
CapsuleVOS~\cite{duarte2019capsulevos} &\XSolidBrush &\color{blue}\textbf{62.3}  &\color{red}\textbf{67.3} &\color{blue}\textbf{68.1} &\color{blue}\textbf{53.7} &\color{blue}\textbf{59.9}\\
\textbf{DTMNet}   &\XSolidBrush &\color{red}\textbf{65.6} &\color{blue}\textbf{66.1} &\color{red}\textbf{68.9} &\color{red}\textbf{60.5}  &\color{red}\textbf{66.8}\\
\hline
\end{tabular}
\end{center}
\label{table4}
\end{table}
\textbf{Results on YouTube-VOS.}
The YouTube-VOS computes $\mathcal{J}$ and $\mathcal{F}$ on seen and unseen categories, denoted as $\mathcal{J}_s$, $\mathcal{F}_s$, $\mathcal{J}_u$, $\mathcal{F}_u$ in Table~\ref{table4}.
The seen categories are included in both the training and the validation sets while the unseen categories only exist in the validation set.
As listed by Table~\ref{table4}, our DTMNet achieves the best global mean $\mathcal{G}$ of $65.6\%$, outperforming the second best-performing CapsuleVOS ($\mathcal{G}=62.3\%$) by a large margin.
Besides, our DTMNet even outperforms the best-performing offline method S2S by $1.2\%$ in terms of $\mathcal{G}$.
Especially, our DTMNet achieves excellent performance on the unseen categories with $\mathcal{J}_u=60.5\%$ and $\mathcal{F}_u=66.8\%$, significantly outperforming the second-best method CapsuleVOS by $6.8\%$ and $6.9\%$ and even outperforming the best OL method S2S by $5.0\%$ and $5.6\%$, respectively. The experimental results demonstrate the favorable generalization capability of our DTMNet to unseen categories. We argue that this is due to the fact that the short-term memory sub-network learning is guided by the semi-supervised loss $\mathcal{L}_{sem}$ (\ref{eq:semi-loss}).
%
%
%
\begin{figure*}[t]
\begin{center}
\begin{tabular}{c}
\includegraphics[width=0.94\linewidth]{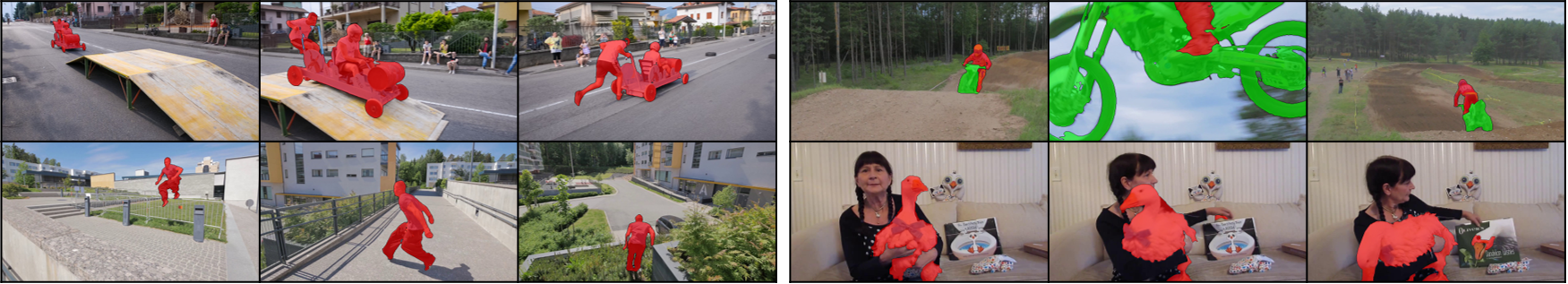}
\end{tabular}
\end{center}
   \caption{Some qualitative results of our DTMNet on DAVIS 2016 val (the first column), DAVIS 2017 val (top of the second column) and YouTube-VOS (bottom of the second column) respectively. The sequences are \textit{soapbox}, \textit{parkour}, \textit{motocross-jump} and \textit{3e03f623bb}. Best viewed in color.}
\label{fig:Comparison}
\end{figure*}
\subsection{Ablation Study}\label{sec:ablative}
We compare three variants of our DTMNet, including those without long-term memory sub-net (DTMNet-L), short-term memory sub-net (DTMNet-S) and graph learning model (DTMNet-W denotes removing the weights in (\ref{eq:Aij})).
We evaluate them on the DAVIS 2016 val and list their results in Table~\ref{table5}.
The DTMNet-S achieves a $\mathcal{J}$ of $81.5\%$, which is lower than the DTMNet by $4.4\%$, which verifies the effectiveness of the short-term temporal information that can help to boost the accuracy of VOS.
Moreover, the DTMNet-L only has a $\mathcal{J}$ of $71\%$, which is significantly lower than the DTMNet by $14.9\%$.
This shows the key role of the long-term temporal information that makes the model robust against occlusions and drifting, which significantly affects the performance of our model.
Finally, we can observe that $\mathcal{J}$ is dropped from $85.9\%$ to $85.2\%$ when removing the weights of the adjacency matrix in (\ref{eq:Aij}), which verifies the effectiveness of using graph learning structure that can also help to boost the performance of our model to some extent.
\subsection{Qualitative Results}\label{sec:qualitative Results}
Figure~\ref{fig:Comparison} shows some qualitatively visual results on DAVIS 2016, DAVIS 2017 and YouTube-VOS datasets.
We select some challenging videos from these three datasets.
We can observe that our DTMNet still achieves favorable segmentation results when the targets suffer from various challenges like fast motion (the first column top), large-scale variations (the first column bottom and the second column top) and interacting objects (the second column bottom).
%
%
\begin{table}[t]
\small
\caption{Ablative experiments of our DTMNet on DAVIS 2016 val. DTMNet-A, A=S, L, W, denotes the DTMNet without short-memory, long-memory and graph learning modules, respectively.}
\setlength{\tabcolsep}{1.4mm}
\begin{center}
\begin{tabular}{|c||c|c|c|c|c}
\hline

Metric   &DTMNet &DTMNet-S &DTMNet-L &DTMNet-W  \\
\hline
$\mathcal{J}$ &85.9 &81.5 &71 &85.2 \\
\hline
\end{tabular}
\end{center}
\label{table5}
\end{table}
\section{Conclusions}
In this paper, we have proposed an end-to-end DTMNnet for VOS which mainly includes a short-term and a long-term memory sub-networks. The former models the fine-grained spatial-temporal interactions between local regions across neighboring frames via a graph-based learning framework, which can well preserve the visual consistency of local regions over time. The latter models the long-range dynamics of object via an S-GRU, making the segmentation robust against occlusions and drift errors. Extensive evaluations on three benchmark datasets including DAVIS 2016, DAVIS 2017 and YouTube-VOS demonstrate favorable performance of our method over state-of-the-art methods in terms of both speed and accuracy.
{\small
\bibliographystyle{ieee_fullname}
\bibliography{egbib}
}
\end{document}